# Pathway toward prior knowledge-integrated machine learning in engineering

Xia Chen and Philipp Geyer
Leibniz University Hannover, Institute for Design and Construction, Sustainable Building Systems Group, Hannover, 30419, Germany

**Abstract**

Despite the digitalization trend and data volume surge, first-principles models (also known as logic-driven, physics-based, rule-based, or knowledge-based models) and data-driven approaches have existed in parallel, mirroring the ongoing AI debate on symbolism versus connectionism. Research for process development to integrate both sides to transfer and utilize domain knowledge in the data-driven process is rare. This study emphasizes efforts and prevailing trends to integrate multidisciplinary domain professions into machine acknowledgeable, data-driven processes in a two-fold organization: examining information uncertainty sources in knowledge representation and exploring knowledge decomposition with a three-tier knowledge-integrated machine learning paradigm. This approach balances holist and reductionist perspectives in the engineering domain.

**Highlights**

- A systematic review of philosophical mindset in current methodologies: reductionism and holism.
- No Free Lunch (NFL) in knowledge representation and problem formalization leads to performance gaps and uncertainties in the building engineering domain.
- Knowledge decomposition paves the path toward knowledge-integrated machine learning - a three-level ladder of integration paradigms.
- Reconciling holism and reductionism methods contributes to effective engineering solutions.

**Introduction**

Modeling, forecasting, and optimizing engineering scenarios as inverse problems with hidden physics are often effort-wisely expensive and require different first-principles or symbolism formulations (Karniadakis et al. 2021). Meanwhile, rapid advancements in artificial intelligence (AI) have attracted attention across a variety of fields. However, significant progress has been observed in areas where the data is fundamental and advantageous, but progress is slower in domains reliant on explicit laws and prior knowledge, such as in building engineering.

First-principles models, rooted in a reductionism philosophy (Andersen 2001), logically decompose, abstract, and deduce the underlying principles of a given phenomenon to formulate symbol-based rules (Minsky 1991) as a concise and abstract representation. These rules hold strong validity when it comes to extrapolation problems and scenario generalizations, hence naturally fit for aligning engineering principles and conducting validation when exploring the design process, e.g., building design with structural engineering. However, these pre-defined, symbolic-based rigid clarity limits their ability to handle cases that lack information definition context or do not fit the rigid rule constraints. Such inflexible architectures and organizational limitations become increasingly pronounced in the face of objectives in multidisciplinary challenges that demand more comprehensive and nuanced definitions of system modeling, such as sustainability (Westermann and Evins 2019). The advantages of pre-defined, context-based rules transform into significant drawbacks of performing efficient searching, manipulating, and validating elements in complex situations, highlighting the need for a more flexible and adaptable modeling approach.

Machine learning (ML) methods, currently dominated by a connectionist approach (Elman 2005), have succeeded in various data-rich fields (L'heureux et al. 2017). These models primarily rely on heuristic connections to learn the mapping between data inputs and outputs, making them broadly applicable through the universal approximation, end-to-end behavior. However, their generalized approach also leads to heavy data reliance and data-hungry issues, as the model organization and approaching mechanisms are not specifically tailored based on the domain data characteristic, especially in engineering fields, where the data is at a certain level interpretable by domain principles and rules. Much of the information is often in the form of expert knowledge or experience (Kasabov 1996). Without the support of such information, obtaining adequate data through measurement, collection, or generation is normally expensive in time and effort, or even impossible. Additionally, the "patterns" learned by ML methods are essentially a result of interpolation in the data space. They do not reflect the underlying domain knowledge but only an emergent byproduct (Hasson et al. 2020). With the risk of biased data sampling, the conclusion might fall as spurious (Chen et al. 2022a). Unlike first-principles methods using symbolic abstraction and expressiveness which are effective in data utilization with extrapolation validity, most domain knowledge is less directly applicable to ML methods.

The limitations previously mentioned in the data-driven process are rooted in the need for a deeper understanding of domain ontology (Gruber 1995). This refers to the formal description of knowledge as a set of concepts within a given domain and their relationships. In light of these limitations, we propose a path toward knowledge-integrated machine learning and argue for its necessity based on the following two critical points:

1. All forms of representation, whether logic-based reductionism or data-driven connectionism, come with inherent cognitive biases. There is often a focus on validating diverse algorithms, but less attention is paid to the diversity in how problems are represented.
2. Imperfect data, difficulty in modeling physics-inversed problems, and limited data with high-dimensional problems cause uncertainty and performance gaps that exist widely in engineering. A parallel utilization of both approaches makes better use of limited resources and data for efficient search and optimization.

The novelty of this study is organized in two parts: the characteristics of knowledge-based and data-driven approaches are identified and analyzed through analogy studies. The insights are then mapped to the existing gap and uncertainty within our domain; Secondly, a systematic discussion is presented on the synergistic knowledge integration across the different stages of the data-driven process based on their natural-inherited characteristics.

## Problem representation: the information uncertainty and performance gap

In the building engineering domain, research on the performance gap (Boghetti et al. 2019; Wilde 2014) has identified differences between: (1) first-principle predictions and measurements; (2) ML and measurements, and (3) measurements from different sources. The first two points are closely linked to the No Free Lunch (NFL) theorem (Wolpert and Macready 1997), originating from optimization problems, which asserts that no single algorithm can outperform all others across all scenarios. We have observed that despite the broad research interest in addressing the gap in the first two points, the third point statement is equally or even more critical for approaching the core of the gap issue because it emphasizes the way of information representation: How is the problem formalized, which knowledge is represented, and how fuzziness of information inherently manifests in data, knowledge-based formalizations, and ML algorithms. Hence, we state that: **there is no one best way to represent knowledge or formalize information for problems.** To fully comprehend the limitations of all these factors, we investigate them from the perspective of uncertainty decomposition and the performance gap presented in engineering domains.

**Uncertainty due to the data**

Consistent with the established statistical tradition, the existence of uncertainty in imperfect data has long been linked to standard probability and probabilistic predictions. These predictions encompass two distinct forms of uncertainty: *aleatory* and *epistemic* (Der Kiureghian and Ditlevsen 2009). Aleatoric uncertainties, arising from the inherent randomness of the natural data generation process, are generally less amenable to reduction. On the other hand, epistemic uncertainties, arising from limited data or understanding, can potentially be mitigated through additional information acquisition or model refinement. Epistemic uncertainties are further decomposable into two subcategories: *parametric* and *structural*. In this context, we address these subcategories individually by assigning them to knowledge-based, first-principles methods and data-driven, ML models based on their respective characteristics. An illustrative summary of this approach is presented in Figure 1.

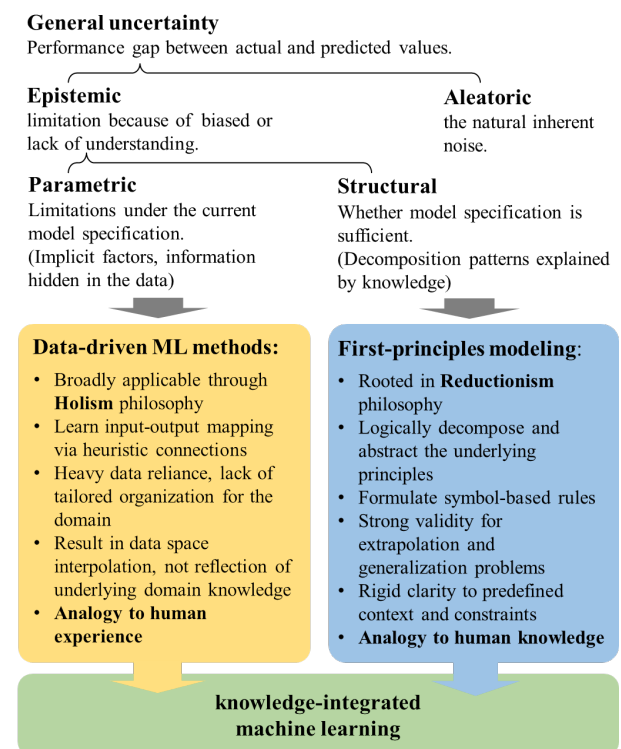

*Figure 1: Uncertainty (performance gap) decomposition and method coverages. Both first-principles and ML methods are complementary advantageous at addressing structural and parametric uncertainties, individually, which forms the theoretical foundation of knowledge-integrated machine learning.*

**Uncertainty due to physics (domain knowledge), symbolism**

Rooted in the reductionism mindset, the first-principles simulation driven by domain knowledge is based on the idea that breaking down a complex problem into smaller parts will necessarily result in a complete understanding of the issue (Sarkar 1992). In this context, domain knowledge comprises static, deterministic rules presented in symbolic formalizations and numerical simulations. Essentially, it is a highly informative abstraction that carries domain expert knowledge within set scenarios, which means it supports more logical computationalism, such as reasoning (Pearl and others 2000) and extrapolation (Balestriero et al. 2021) if the scenario follows the predefined principles and rules, this refers to system 2 (Bengio et al. 2021). However, this decomposable mindset has a potential fallout (Hasson et al. 2020; Pigliucci 2014) since complex systems can exhibit emergent properties that cannot be predicted from the properties of the individual components alone, as stated in the old dictum: '*The whole is more than the sum of its parts*' (Becht 1974). Normally, this limitation is caused by biased or limited understanding of a system, corresponding to the definition of epistemic structural uncertainty (Jeremiah Liu et al. 2019). In complex systems with high-dimensional problem space (Smith-Miles et al. 2014), we summarize the following sources of biases and performance gaps by utilizing first-principles modeling:

- **Model over-simplification**: An over-simplification of complex systems leads to capturing the nature of the system less accurately with the lack of description regarding weak interactions or dynamics in the system. This over-simplification is due to the missing information to conduct prior knowledge or drawbacks in computational efforts. For example, a model that assumes linear relationships between variables may not be appropriate for non-linear systems: In building simulation of structure engineering, a linear model that assumes a simple relationship between a component's load and its deformation may be insufficient for accurately predicting the behavior of the structure under extreme conditions (Stochino 2016).
- **Confirmation bias in modeling**: The knowledge in the defined context describes only a limited set of scenarios or processes in a system and ignores other important factors. This reliance on prior knowledge can lead to confirmation bias, resulting in a selection of first-principles models that are based on biased or incomplete assumptions, which has been confirmed in the engineering design community (Hallihan and Shu 2013) that leads to suboptimal reasoning. In other words, the reliance on informative priors (Karniadakis et al. 2021) does not guarantee inferential perfection or even consistency in problem-solving (Minsky 1991). For example, we point to a study in energy system optimization modeling (DeCarolis et al. 2017), which suggest that the model results must be synthesized into insights by combining, e.g., spatiotemporal boundaries, before presentation to decision-makers because general caveats may be ignored by supporters, or used by critics to challenge the analysis.
- **Context constraints in model development**: First-principles methods usually rely on rigid adherence to predefined context and symbol-based rules derived from a strict logical deduction process. This process requires dedicated data preparation and meticulous modeling for a case-specific context, which can limit the ability to accommodate exceptional conditions and implicit interactions, as the model architecture is not flexible enough to process the system's behavior completely. We found numerous references highlighting this constraint when transitioning from experimental modeling or simulation in lab environments to real-world projects, such as integration between BIM, IoT (Tang et al. 2019), and implementation in building energy assessment among different simulation tools and regions (Durdyev et al. 2021).

**Uncertainty due to the learning models (ML), connectionism**

The data-driven methods process the problem in a "direct-fit" (Hasson et al. 2020) and holistic manner. This approach treats the pattern-fitting problem as a whole and consists of designing three main components: the *model architecture* (how is it organized?), the *learning rules* (how does it learn?), and the *objective functions* (what does it learn?) (Richards et al. 2019). Data-driven methods provide a unique approach to problem-solving. They serve a complementary role to symbolism approaches, particularly in avoiding cognitive bias and constraints. However, these methods have their own gaps, which we have organized into three categories. Notably, these gaps can potentially be mitigated in engineering scenarios through the integration of prior domain knowledge:

- **Approximation error:** Whether the ML model organization (e.g., the design of the model structure, depth of model) approximates a solution to accurately describe complex system behavior. For example, in building energy performance forecasting, whether a model is designed to capture the autocorrelation (day hours), long-term patterns (seasons), and short-term variation (user-occupancy changes) is essential.
- **Optimization error:** The choice of learning rules, such as gradient descent, Hebbian (Movellan 1991), or backpropagation (LeCun et al. 1988), also impacts the model's accuracy in different scenarios. If the learning rule is not fitting for the task optimization or the data format, it may cause difficulty in finding or result in convergence to a suboptimal solution. This difficulty also refers to the over-/underfitting issues in the model training process (Dietterich 1995).
- **Generalization error:** The choice or the design of objective function, such as mean squared error or cross-entropy (Shore and Johnson 1981), is applied to measure the difference between the prediction and the ground truth value. This error also refers to the issue of whether the training error minimization to approaching the defined indicator leads to a more accurate prediction for the solution, and how well it fits other tasks.

The data-driven process, more analog to experience (probabilistic, variations, statistic), is strength in interpolation, implicit pattern finding, and predictive control based on the information feed-in via data and map to the task in an end-to-end behavior, which refers to system 1 (Bengio et al. 2021), and efforts to specify the process contribute to reducing parametric epistemic uncertainty in engineering problems.

## The Ladder of knowledge-integrated machine learning

Once we acknowledge the characteristic of all elements and linked to the decomposed uncertainties in the performance gap, we organize a systematic integration pathway in a three-level manner, the ladder of knowledge integration, as presented in Figure 2:

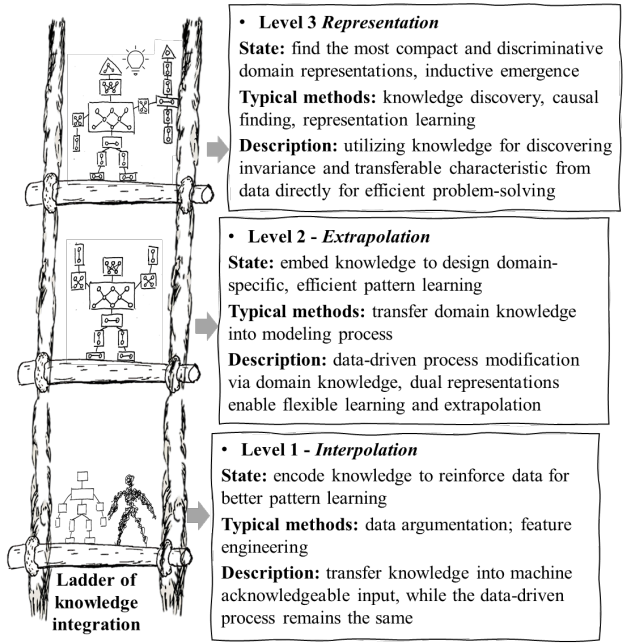

*Figure 2: The ladder of knowledge-integrated machine learning. The three levels in the pathway state their difference and core ability, linking to their typical methods and characteristic descriptions. A higher level owns and is compatible with lowers abilities.*

The "Ladder of Knowledge Integration" is designed to bridge and complement the advantages of modeling in connectionism (data-driven methods) and symbolism (knowledge-based methods). Connectionism excels in its heuristic nature but is weak in induction, whereas symbolism behaves oppositely. The level of the ladder presents the knowledge integration degrees into the data-driven process and what abilities the combined method achieves. We aim to construct a pathway that embodies the heuristic connections between knowledge.

### Knowledge-based decomposition

Before we step on the ladder, we must revisit our embodied prior knowledge from the perspective of decomposition. For a machine to learn, it must have a means of representing the knowledge it will acquire. This takes the form of a mechanism, data structure, or representation that embodies knowledge. In engineering, we've identified several ways of decomposing systems or framing problems. These include:

- **Domain decomposition:** use domain knowledge to segment a task and describe in subfactors that are independent with formulars, e.g., in building performance forecasting, we separate building energy consumption to heating/cooling load modeling, or separate to external/internal factors modeling.
- **Mathematical decomposition:** much of the data collected from systems carry information in a compact, implicit behavior. In this context, scientific or mathematical processing is a knowledge-embedded way widely existed in solving engineering tasks applied for extracting valuable information or denoising, e.g., information extraction of a system time-series data using Fourier transformation, STL decomposition (Chen et al. 2022b; Cleveland et al. 1990), wavelet decomposition (Torrence and Compo 1998), data statistical characteristics extraction (Lubba et al. 2019; Montero-Manso et al. 2020), etc.
- **Complexity decomposition:** Describing a system at multi-scales, corresponds to a focus on capturing the behavior at different levels of complexity and features for better understanding the system behavior. For example, describing building energy performance at different scales and aggregating: the building level with energy system dimensioning, zone-level with occupant behavior analysis, and building component level with thermal behavior (Geyer et al. 2021) and structure modeling (Geyer 2009).

To better illustrate different knowledge decomposition concepts, we demonstrate an example in the context of building energy demand forecasting in Figure 3.

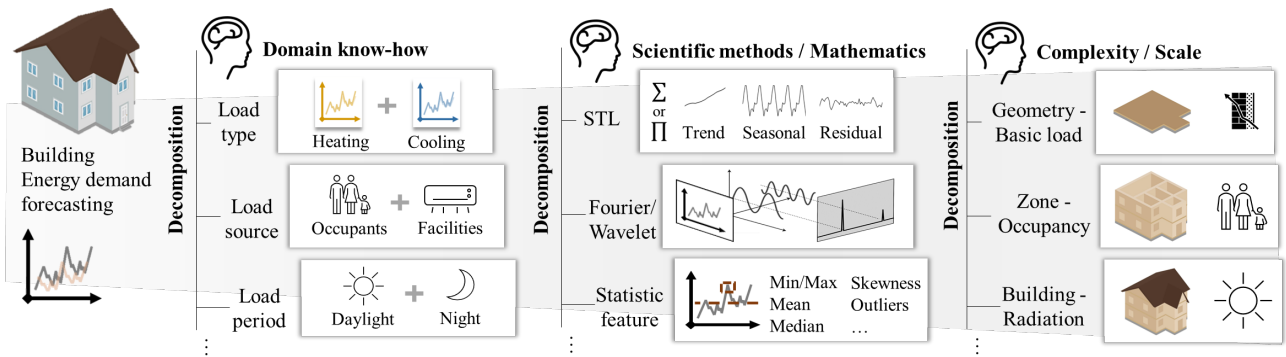

*Figure 3: Knowledge-based decomposition in an example of building energy demand forecasting task. Knowledge, including domain know-how, mathematical tools, and different system scales/complexities is embedded to help decompose the energy demand time series and gain more information for solving the problem.*

Essentially, the purpose of knowledge decomposition and data processing is to transfer information into a unified, machine-learnable representation, which requires two-side extending efforts: on the knowledge formulation, research in more efficiently exploit informative abstraction as domain knowledge for representation; on the data-driven modeling, design ML methods that learn various ways to represent knowledge; Corresponding to the structure presented in Figure 2, Figure 4 demonstrates methodology-wised three levels of knowledge-integrated machine learning.

**Level 1: Interpolation**

The contribution at Level 1 focuses on data augmentation for the input: formalizing the domain knowledge into feature engineering or using first-principles simulation to generate synthetic data, and feed-in into ML methods. Various patterns of abstraction via the first-principles model as extra data input to reinforce the training process (via super-/unsupervised learning) or constructing environment (via reinforcement learning). The synthetic data is a representation form to carry extra informative input for the ML modeling process, which serves for observational data biases reduction, implicit pattern reinforcement, and multi-scale information input, such as overall system indicators (e.g., the degree of system chaos, coherence, or other statistic indicators).

At this level, the primary efforts for integrating knowledge come down to reinforcing additional information for the available data. The new combined model possesses advantageous characteristics in solving engineering tasks: Due to the logical and rule-based nature, domain knowledge fits in interpreting different scenarios and generates corresponding supportive data through inference, thereby improving the performance of the data-driven process. Furthermore, it is possible to integrate domain knowledge with data argumentation in an incremental manner (Castro et al. 2018), where the additional knowledge input further enriches the information content of the data, tuning the input through the feedback of the model's output, which is also referred to as feature engineering. From this perspective, domain knowledge contributes to the model explainability in understanding the input feature importance and output results (Roscher et al. 2020). As an example, we point to a hybrid-model framework in building load forecasting tasks by blending different detail-level building simulation data with real-world records for ML training to minimize the performance gap while alleviating the modeling efforts (Chen et al. 2022b).

Despite the advantages that Level 1 carries, the data-driven process itself remains unchanged in terms of efficiency. Thus, while the combined process alleviates the data-hungry issue at a certain level and hence is better suited to solving problems in niche domains, the underlying learning mechanism of the model remains an interpolative behavior in high-dimension problem space (Smith-Miles et al. 2014). In other words, the model's credibility is still limited within the input/training data ranges, and the same ML model's interpretability and transparency level as the black-box behavior.

**Level 2: Extrapolation**

At this level, domain knowledge is integrated into the data-driven process itself, providing constraints or specifications on the objective function, learning rules, or model architecture of ML models based on the task. The modified model learns the data within a specific domain more efficiently and is capable of extrapolation in boundary value problems at a certain level due to the symbolic and logical capabilities of integrated knowledge.

For the objective function, which defines what a good prediction is and is not, it offers a nexus of incorporating information from measurement data and embedding knowledge through equations via objective function specification. For instance, in the case of the Physical-Informed Neural Network (PINN) (Karniadakis et al. 2021), partial differential equations (PDEs) are incorporated into the loss function of a neural network using automatic differentiation. The network is then trained to determine the optimal parameters by minimizing the combined loss. This combined PDEs and data-driven approach is conveniently applied to describe various physical processes, e.g., heat transfer and indoor environment simulation in building engineering (Drgoňa et al. 2021) .

Regarding the learning rule, the integration of knowledge provides information on additional regularization or constraints, allowing the data-driven process to avoid falling into invalid parameter space and to learn the mapping between inputs and outputs more efficiently. For instance, B-PINN (Yang et al. 2021) builds upon PINN by introducing a Bayesian framework into the data-driven process, transforming the learning process into an update of prior beliefs to filter out the noise data from the input, thereby enhancing the robustness of the model.

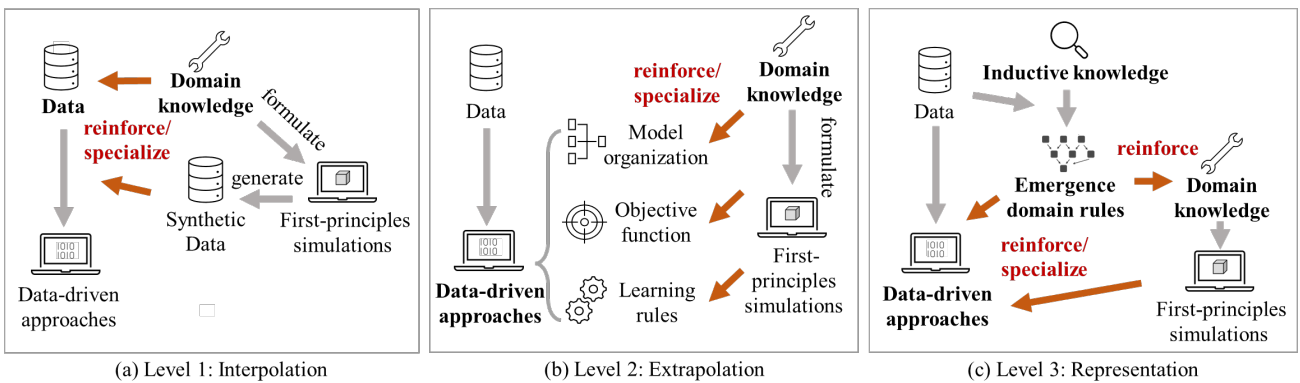

(a) Level 1: Interpolation  (b) Level 2: Extrapolation  (c) Level 3: Representation

*Figure 4: Process illustrations of knowledge-integrated machine learning at three levels in the ladder of knowledge integration.*

In terms of the model organization, the integration incorporates the decomposed information from the prior domain knowledge into the organization. For example, component-based machine learning (CBML) (Geyer and Singaravel 2018) formalizes the topological structure information of the building and training corresponding ML model sets at different scales: components, zones, and buildings. The components and zones are reusable to reassemble and predict the performance of new building structures, achieving a better generalization performance, or in a way, extrapolating (Singaravel et al. 2019). Additionally, this modular mindset of the model organization also enhances the interpretability and transparency of the building modeling since the error is trackable on a component basis (Geyer et al. 2021; Singaravel et al. 2020). Furthermore, the component-based training approach enables the model to effectively extract information when the available data is small, fragmented, imperfect, or biased (Chen et al. 2023).

At this level, the most notable feature is that the ML modeling process is not solely data-driven. The reconstruction and integration of knowledge serve as a "skeleton" that allows the model modification to be more specialized in the corresponding engineering domain, combining the expressiveness and procedural versatility of symbolic representations with the fuzziness and adaptiveness of data representations, while reconciling the model data-hungry issue. Compared to Level 1, Level 2 possesses a certain level of interpretable and extrapolation capability.

### Level 3: Representation

As a short recap of the two levels mentioned above: In Level 1, the data input is reinforced or specialized through domain knowledge, thus improving the interpolative performance of the ML model; while in Level 2, the domain knowledge is integrated into the data-driven process, endowing a better interpretable modeling process with a procedural logic, and extrapolative capabilities. It is worth noting that the "domain knowledge" in these levels' context is referred to as *first-order knowledge*, as *"learn to do"*, which is guided direct resolution of domain problems through integration with the data-driven process. This type of knowledge primarily originates from human input. Although it has been widely proved that the combination certainly boosts the enhancement of accuracy and interpretability in the data-driven model, but also carries the risk of lacking feedback cognitive biases in the input knowledge (Minsky 1991).

At Level 3, the objective is to enable the ML process to understand, extract information from engineering problems through data, and model it in an unsupervised or semi-supervised manner, which requires the knowledge for integration as *second-order knowledge*, that is, *"learn to learn"*, so that enables the modeling process to discover and induce invariance and transferable information patterns from the data. Transferring such patterns into machine acknowledgeable representations corresponds to the first-order knowledge in previous levels. In other words, combining second-order knowledge in the process helps to discover domain rules from the data in an unsupervised behavior, reducing the risk of prior knowledge biases, hence encoding, representing, and transforming effective information concisely and self-continuously from domain data. Such second-order knowledge includes, but is not limited to:

- **Causal dependencies asymmetry**: Causal inference (Pearl and others 2000) for determining causation relationships between different variables and output, conducting reasoning to answer "what-if" questions via intervention or counterfactual analysis.
- **Entropy reduction**: Self-supervised learning (Jing et al. 2022), Supervised contrastive learning (Khosla et al. 2020), goodness (Hinton 2022), rate reduction (Ma et al. 2022) to distinguish good and bad for representation learning.
- **Occam's razor**: Representation compression via manifold learning (Lin and Zha 2008), Autoencoder (Ng and others 2011), VAE (Kingma et al. 2019), etc, to embody effective information in a more compact format.

Moreover, the emerged domain rules by second-order knowledge are able to be obtained from different data sources, accumulated, and transferred, such as causal dependencies (Judea 2010; Schölkopf 2019). For instance, in building performance modeling, the causal inference helps to establish a causal skeleton between variables and avoid biased, spurious conclusions by data-driven process (correlation does not imply causality: a lower building structure strength doesn't lead to better energy performance building but because of smaller geometry and higher compactness) (Chen et al. 2022a). As the building model is further refined, the variables and relationships in the causal skeleton are adaptative to be expanded, transferred, modified, and applied to other building simulation projects. Meanwhile, the ML models that incorporate causal discovery own the capability to answer "what-if" questions through intervention and make unbiased inferences within the limited data, here, the representation is the causal dependencies among variables.

### Discussion

In engineering, we tend to use comprehensive, human-interpretable criteria to describe, explain, and model a system or phenomenon, which accounts for dominance in most of history, and name it with scientific principles. In reality, we noticed that the gap between prediction value and measurement widely exists with the in-parallel running first-principles methods and ML methods, as well as the uncertainty arising from data information collection and knowledge formalization. Yet, we only pay less attention to evaluating their limitations originated by their rooted reductionist and holism philosophy. Therefore, solving the problem of describing and solving complex systems requires more than just a single level of perspective. Currently, the promising performance of data-driven methods and direct-fit information input from data, characterized by end-to-end behavior, has reshaped our definition of "understanding". We believe that reevaluating and embracing holistic thinking will enlighten a vital path to bridging performance gaps in the engineering field.

## Conclusion

In this study, we proposed a systematic pathway to integrate multidisciplinary domain professions into machine-acknowledgeable, data-driven processes. Our investigation initially focused on understanding their rooted philosophy, the gap in problem representation, and uncertainty decomposition in engineering domains. Based on our findings, we summarized a three-level paradigm of a knowledge-integrated machine learning ladder to effectively utilize the knowledge and data-driven approaches in applying the building engineering domain. As a direct result of our study, we have set the groundwork for effective data-driven engineering design and decision-making support. The pathway underscores the potential and applicability of the approach in enhancing the efficacy of engineering solutions.

## Acknowledgement

We gratefully acknowledge the German Research Foundation (DFG) support for funding the project under grant GE 1652/3-2 in the Researcher Unit FOR 2363 and under grant GE 1652/4-1 as a Heisenberg professorship.